\documentclass[letterpaper, 10 pt, conference]{ieeeconf}  

\IEEEoverridecommandlockouts                              

\usepackage{graphics} 
\usepackage{graphicx}
\usepackage{graphicx}
\usepackage{epsfig} 
\usepackage{mathptmx} 
\usepackage{times} 
\usepackage{amsmath} 
\usepackage{amssymb}  
\usepackage{amsmath}
\usepackage{booktabs}
\usepackage{microtype}

\usepackage{amsmath,amsfonts}
\usepackage{algorithmic}
\usepackage{algorithm}
\usepackage{float}
\title{\LARGE \bf
Leg-Arm Coordinated Operation for Curtain Wall Installation
}

\author{Xiao Liu$^{1}$, Weijun Wang$^{1}$, Tianlun Huang$^{1}$, Zhiyong Wang$^{1}$,  Wei Feng $^{1,2*}$
\thanks{*corresponding author. e-mail: wei.feng@siat.ac.cn}
\thanks{$^{1}$ All authors are with Shenzhen Institute of Advanced Technology, Chinese Academy of Sciences, Shenzhen, 51800, China.
        {\tt\small  Contact: xiao.liu1@siat.ac.cn}}%
 \thanks{$^{2}$ All authors are with Shenzhen University of Advanced Technology and University of Chinese Academy of Sciences, Shenzhen, 51800, China.
        {\tt\small  Contact: wei.feng@siat.ac.cn}}%
}

\begin{document}

\maketitle
\thispagestyle{empty}
\pagestyle{empty}

\begin{abstract}

With the acceleration of urbanization, the number of high-rise buildings and large public facilities is increasing, making curtain walls an essential component of modern architecture with widespread applications. Traditional curtain wall installation methods face challenges such as variable on-site terrain, high labor intensity, low construction efficiency, and significant safety risks. Large panels often require multiple workers to complete installation. To address these issues, based on a hexapod curtain wall installation robot, we design a hierarchical optimization-based whole-body control framework for coordinated arm-leg planning tailored to three key tasks: wall installation, ceiling installation, and floor laying. This framework integrates the motion of the hexapod legs with the operation of the folding arm and the serial-parallel manipulator. We conduct experiments on the hexapod curtain wall installation robot to validate the proposed control method, demonstrating its capability in performing curtain wall installation tasks. Our results confirm the effectiveness of the hierarchical optimization-based arm-leg coordination framework for the hexapod robot, laying the foundation for its further application in complex construction site environments.

\end{abstract}

\section{INTRODUCTION}
With the acceleration of urbanization, the number of high-rise buildings and large public facilities has steadily increased, making curtain walls—an essential component of modern architecture—widely used [1]. Traditional curtain wall installation is labor-intensive, inefficient, and poses significant safety risks; large panels frequently require multiple workers to install [2]. Construction sites feature diverse terrains, including rugged surfaces, uneven ground, flat areas, narrow passages, and obstacle-dense zones. Currently, curtain wall installation robots are mainly derived from industrial robots and construction machinery. However, industrial robots have low payload-to-weight ratios and are designed for structured environments, whereas construction machinery, despite its high load capacity, suffers from bulkiness, poor flexibility, and low precision, making it unsuitable for precise operations in confined spaces. Therefore, we designed a hexapod curtain wall installation robot.

\begin{figure}[htb]
      \centering
      \includegraphics[width=\columnwidth]{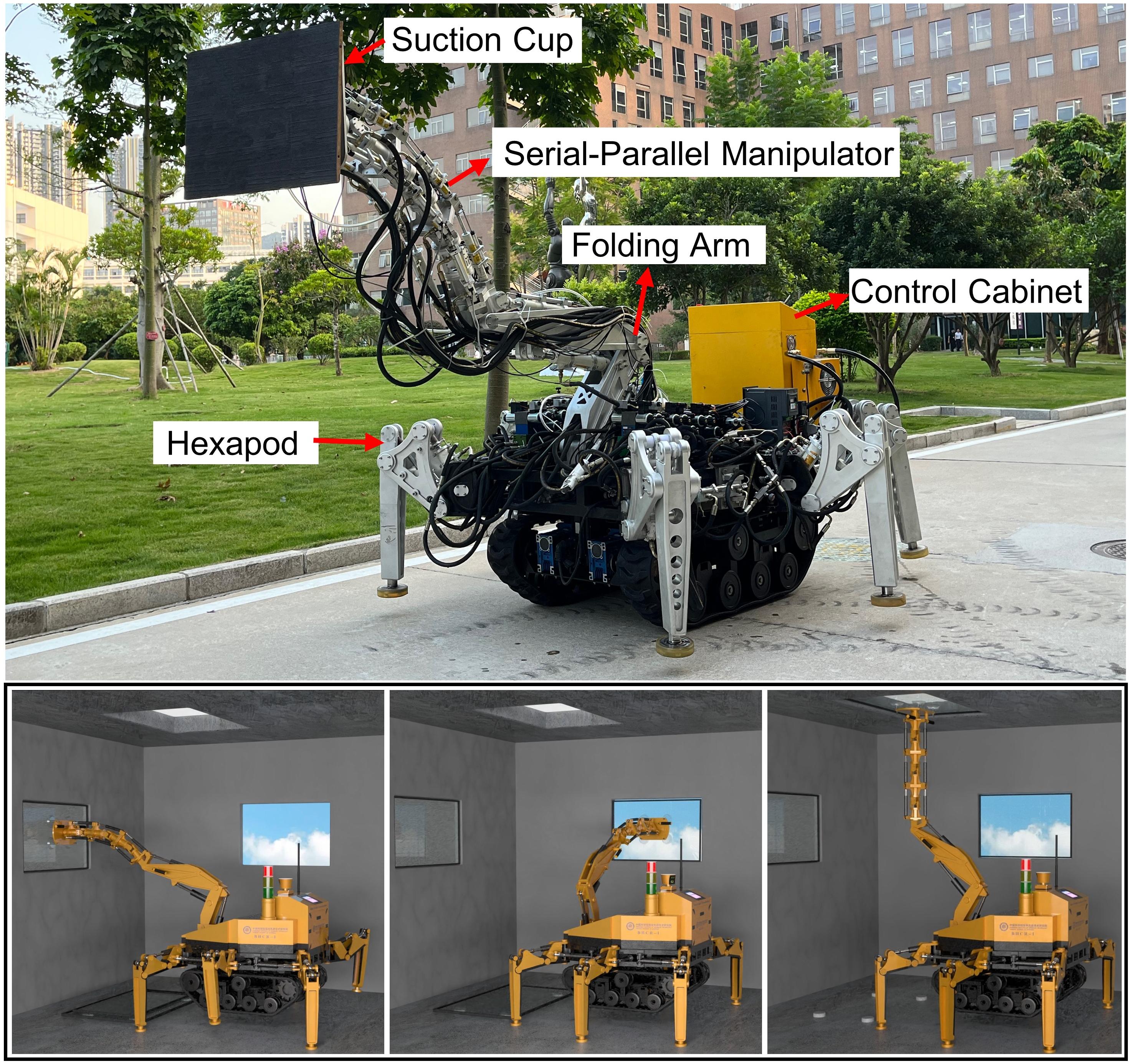}
      \caption{The Curtain Wall Installation Robot}
      \label{figurelabel}
   \end{figure}

Traditional hexapod robots typically focus on mobility while neglecting arm manipulation capabilities. In recent years, leg-arm coordination has become a key technology for enhancing versatility and flexibility [3-10]. [11] introduced an MPC-based Whole Body Control (WBC) strategy for mobile manipulators. [12] developed the SDUHex hexapod robot, featuring six limbs driven by 20 proprioceptive motors. Experiments demonstrated single-arm pushing and dual-arm picking tasks, highlighting the robot’s innovative multiplexing limb functions. A whole-body control method was also tested on sloped terrains and during arm operations, showcasing the limb's multi-functional capabilities.[13] proposed a whole-body control approach for a hybrid serial-parallel humanoid robot. [14] presented an inverse dynamics-based whole-body controller for a torque-controlled quadrupedal manipulator capable of performing locomotion while executing manipulation tasks. In [15], the quadruped robot SDU-ADog was enhanced by integrating a torque-controlled 6-DOF arm, and a novel control framework was proposed that combines virtual model control (VMC) with prioritized whole-body control (WBC). [16] proposed a whole-body planning framework that unifies dynamic locomotion and manipulation tasks by formulating a single multi-contact optimal control problem. [17] presented practical enhancements of the operational space formulation (OSF) to exploit inequality constraints for whole-body control of high-degree-of-freedom robots with a floating base and multiple contacts, such as humanoids. [18] presented a hierarchical model predictive control (MPC) framework for the end-effector tracking control problem of a legged mobile manipulator. To achieve stable movement on uneven terrains for wheeled biped robots (WBR). [19] proposed a whole-body motion control framework based on hierarchical model predictive control (HMPC).

Most current research on leg-arm coordination focuses on quadruped or humanoid robots. Our work aims to achieve leg-arm coordination in a mobile robot featuring a hexapod-based serial-parallel manipulator. As shown in Figure 1, the curtain wall installation robot includes six legs, a folding arm, and a serial-parallel manipulator. By implementing whole-body control on this mobile robot, we achieved coordinated leg-arm control, enabling simultaneous movement and installation. This innovation has significant practical value and represents an important advancement in the field.

\section{A CURTAIN WALL INSTALLATION ROBOT}
The curtain wall installation robot consists of three main components: a mobile chassis, a folding arm, and a serial-parallel manipulator. The folding arm and the serial-parallel manipulator constitute the manipulation part for handling curtain walls, while the mobile chassis provides mobility, as shown in Figure 2. The folding serial-parallel manipulator is 2.3 meters in length and is driven by hydraulic cylinders. The serial-parallel manipulator comprises three redundantly actuated parallel mechanisms, allowing for compliant operation across a vertical range of 150 degrees. The folding arm features three degrees of freedom, comprising a base platform, two support arms, three hydraulic cylinders, and an end-effector platform. It can extend the installation height to 3 meters, further expanding the workspace. Additionally, its foldable design allows access to narrow and indoor spaces. The folding arm meets operational requirements, offering high load capacity and stable performance.

\begin{figure}[htb]
      \centering
      \includegraphics[width=\columnwidth]{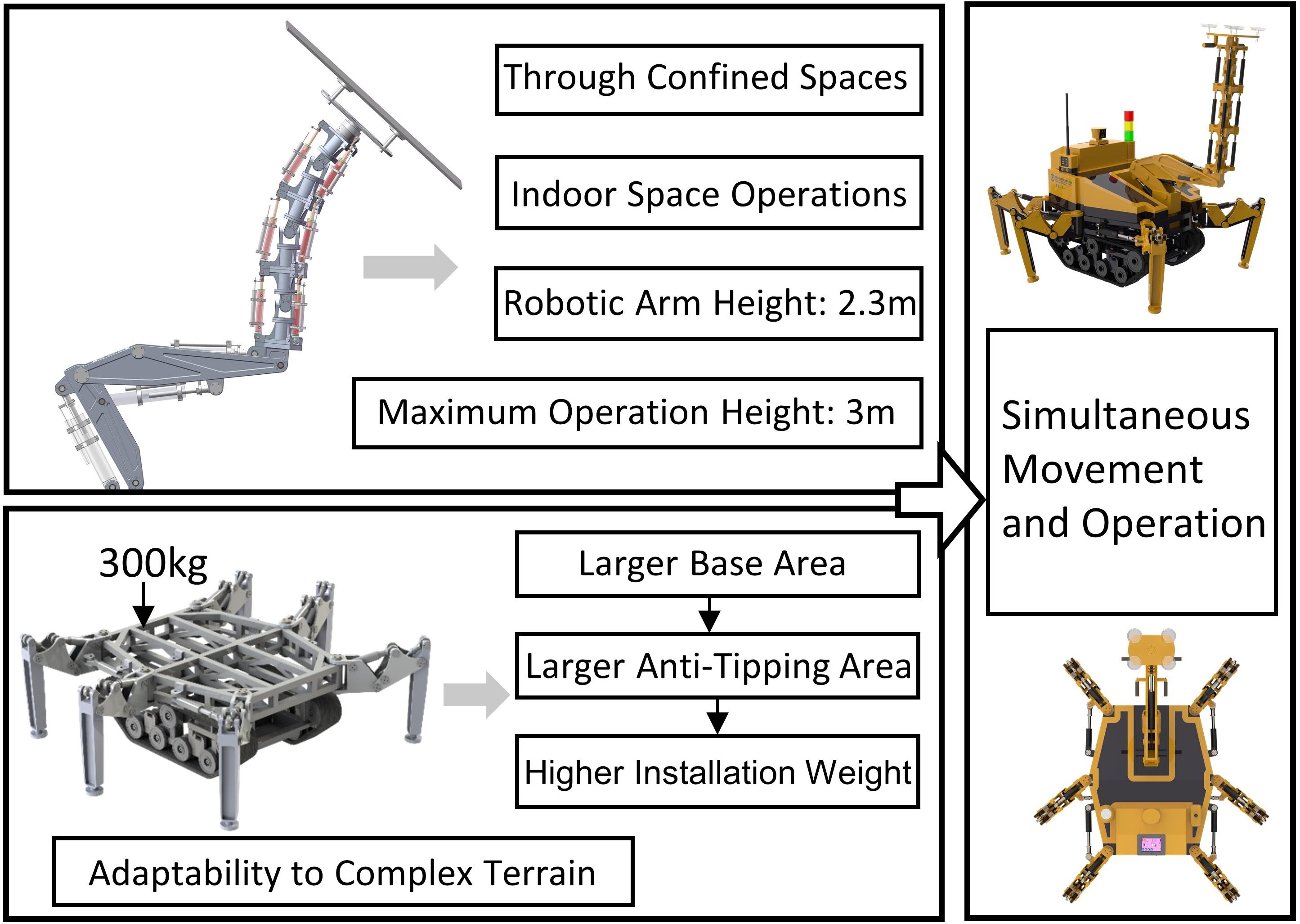}
      \caption{Simultaneous Mobility and the Operation of the Curtain Wall Installation Robot}
      \label{figurelabel}
   \end{figure}

The hexapod mobile chassis can support a payload of 300 kg and is adaptable and flexible on complex terrains, including loose sand, rugged and uneven surfaces, slopes, construction debris, and obstacle-dense environments. Its symmetric polygonal topology design maximizes the chassis area, resulting in a larger anti-tipping area. This improves the stability of coordinated arm-leg operations and sets the stage for subsequent research.

\section{MATHEMATICAL MODEL}
\subsection{Mapping Relationship Between Hydraulic Cylinder Extension and Joint Angles (Manipulator)}
The robotic arm consists of three modules connected in series. We analyze the relationship between the hydraulic cylinder and joint angle changes for a single module. The single module is shown in Figures 3(a) and 3(b). The geometric center point of the zero-position state of the steering joint is chosen as the origin \(O\), and a fixed coordinate system \(\{O\}_{xyz}\) is established. The hydraulic cylinders at points \(P\) and \(Q\) are denoted as hydraulic cylinder \(P\) and hydraulic cylinder \(Q\), respectively. These can be equivalently represented by a virtual hydraulic cylinder at point \(M\) on the \(XZ\) plane, with the same length as the hydraulic cylinders at points \(P\) and \(Q\). Hydraulic cylinder \(KL\) is the primary working cylinder, while hydraulic cylinders \(P\) and \(Q\) are auxiliary working cylinders. During rotation, the extension directions of hydraulic cylinder \(KL\) and hydraulic cylinders \(P\) and \(Q\) are set to be opposite.

The hydraulic cylinder \(KL\) is taken as the subject of study. Let the original length of the hydraulic cylinder be \(l_{KL}\), and its extension be \(\Delta l_{KL}\). The virtual hydraulic cylinder has an original length of \(l_{MR}\) and an extension of \(\Delta l_{MR}\). Point \(S\) is the geometric center of the joint connecting surface \(K'\). A perpendicular line from point \(K\) to line segment \(SO\) intersects at point \(N\).
Assume that after extension, \(SO\) rotates counterclockwise around the center point by \(\theta_4\), moving point \(K\) to \(K'\). Points \(S\) and \(N\) also rotate counterclockwise around point \(O\) by \(\theta_4\) to points \(S'\) and \(N'\), respectively. The virtual hydraulic cylinder's point \(M\) moves to \(M'\).

Based on the geometric structure, it is known that triangles \(\triangle KNO\) and \(\triangle KN'O\) are congruent, and \(\triangle K'OL\) is an isosceles triangle. Given the angles \(\angle KLO\), \(\angle S'OK'\), \(\angle XOL\), and the length \(l_{LO}\) of segment \(LO\), we can derive the following:
\begin{equation}
\label{deqn_ex1}
\angle SOS^{\prime}=\angle KOK^{\prime}=\theta_4
\end{equation}

\begin{equation}
\label{deqn_ex1}
\frac{\sin(\frac{\theta_4}{2}+\angle KLO)}{l_{LO}}=\frac{\sin(\angle KOL-\theta_4)}{l_{K^{\prime}L}}
\end{equation}

The extension \(\Delta l_{KL}\) of hydraulic cylinder \(KL\) is derived as follows:
\begin{equation}
\label{deqn_ex1}
\Delta l_{KL}=l_{KL}-l_{K^{\prime}L}=l_{KL}-\frac{\sin(\angle KOL-\theta_4)\cdot l_{LO}}{\sin(\frac{\theta_4}{2}+\angle KLO)}
\end{equation}

Extension of hydraulic cylinders at points P and Q:
\begin{equation}
\label{deqn_ex1}
\Delta l_P=\Delta l_Q=\Delta l_{MR}=l_{MR}-l_{M^{\prime}R}=l_{MR}-\frac{\sin(\angle MOR+\theta_4)\cdot l_{RO}}{\sin(\angle MRO-\frac{\theta_4}{2})}
\end{equation}
The forces exerted by hydraulic cylinders \(KL\), \(P\), and \(Q\) are denoted as \(F_4\), \(F_5\), and \(F_6\), respectively. The torque \(T_4\) experienced by the module is given by:
\begin{equation}
\label{deqn_ex1}
T_4=\sin\angle K^{\prime}LO\cdot F_4l_{LO}-\sin\angle M^{\prime}RO\cdot(F_5+F_6)\cdot l_{LO}
\end{equation}
Based on the above derivation, the mapping relationship between the hydraulic cylinder extension and joint rotation angles for a single module is obtained, as shown in Figure 3(c):
\begin{figure}[htb]
      \centering
      \includegraphics[width=\columnwidth]{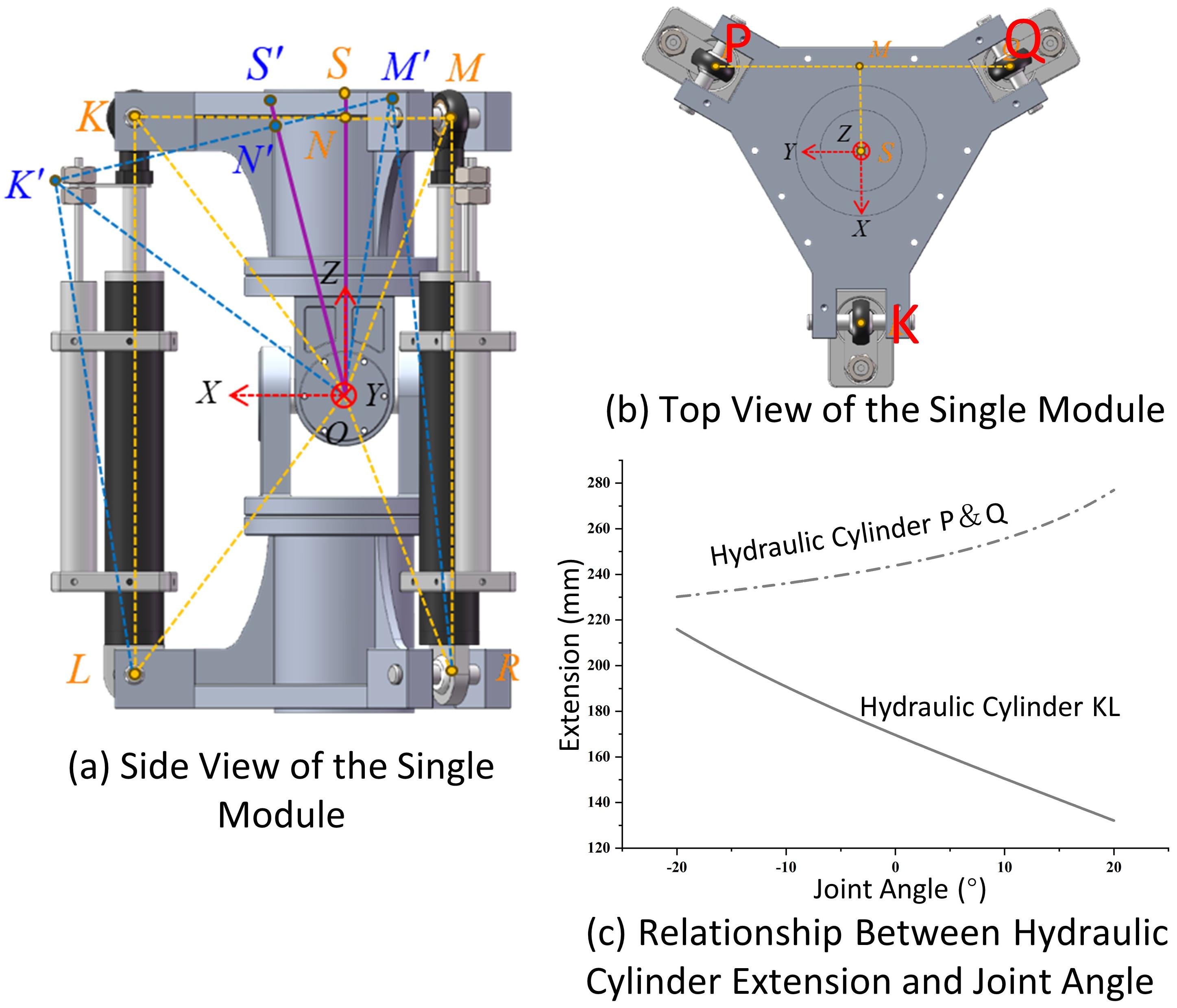}
      \caption{Single Module of the Serial-Parallel Manipulator}
      \label{figurelabel}
   \end{figure}
\subsection{Mapping Relationship Between Hydraulic Cylinder Extension and Joint Angles (Folding Arm)}
The folding arm is composed of three hydraulic cylinders, bearings, and non-standard structural components. The overall dimensions measure 950 mm × 230 mm × 1010 mm, with an end-effector load capacity of 80-100 kg.
\begin{figure}[htb]
      \centering
      \includegraphics[width=0.85\columnwidth]{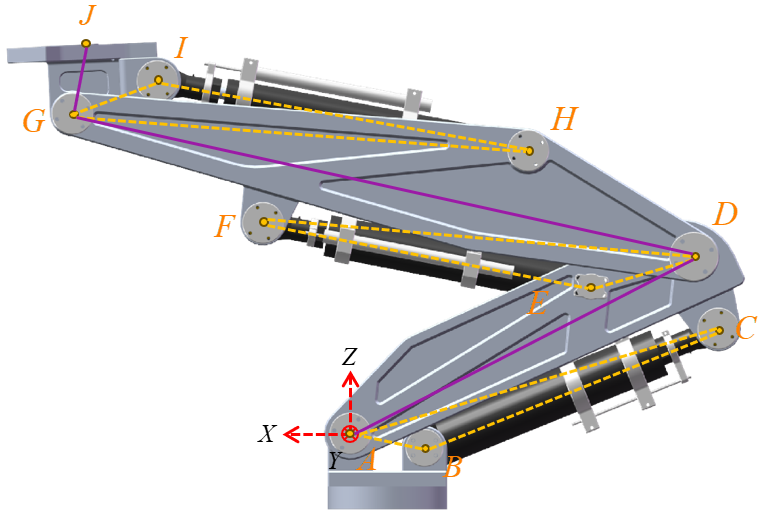}
      \caption{Model of the Folding Arm}
      \label{figurelabel}
   \end{figure}

As shown in Figure 4, let the original lengths of hydraulic cylinders \(BC\), \(EF\), and \(HI\) be \(l_{BC}\), \(l_{EF}\), and \(l_{HI}\), respectively. The corresponding extensions are \(\Delta l_{BC}\), \(\Delta l_{EF}\), and \(\Delta l_{HI}\). The lengths of segments \(AC\), \(DF\), and \(GH\) are known and constant, denoted as \(l_{AC}\), \(l_{DF}\), and \(l_{GH}\), respectively. Similarly, the lengths of segments \(AB\), \(DE\), and \(GI\) are also known and constant, denoted as \(l_{AB}\), \(l_{DE}\), and \(l_{GI}\).

Assume that joint \(DA\) rotates counterclockwise around point \(A\) by \(\theta_1\). The angle \(\angle BAC\) increases by \(\theta_1\) accordingly. Using the cosine rule, we obtain:
\begin{equation}
\label{deqn_ex1}
\Delta l_{BC}=\sqrt{l_{AB}^2+l_{AC}^2-2l_{AB}\cdot l_{AC}\cdot\cos(\theta_1+\angle BAC)}-l_{BC}
\end{equation}
Based on the above geometric derivation and the limitations of the folding arm joints, the mapping relationship between the hydraulic cylinder extension and the angle of joint one can be obtained, as shown in Figure 5(a).
Similarly, assume that joint two \(GD\) rotates clockwise around point \(D\) by \(\theta_2\), and joint three \(JG\) rotates counterclockwise around point \(G\) by \(\theta_3\). It can be deduced that:
\begin{equation}
\label{deqn_ex1}
\Delta l_{EF}=\sqrt{l_{DE}^2+l_{DF}^2-2l_{DE}\cdot l_{DF}\cdot\cos(\theta_2+\angle EDF)}-l_{EF}
\end{equation}
\begin{equation}
\label{deqn_ex1}
\Delta l_{IH}=\sqrt{l_{GI}^2+l_{GH}^2-2l_{GI}\cdot l_{GH}\cdot\cos(\theta_3+\angle IGH)}-l_{IH}
\end{equation}
\begin{figure}[htb]
      \centering
      \includegraphics[width=\columnwidth]{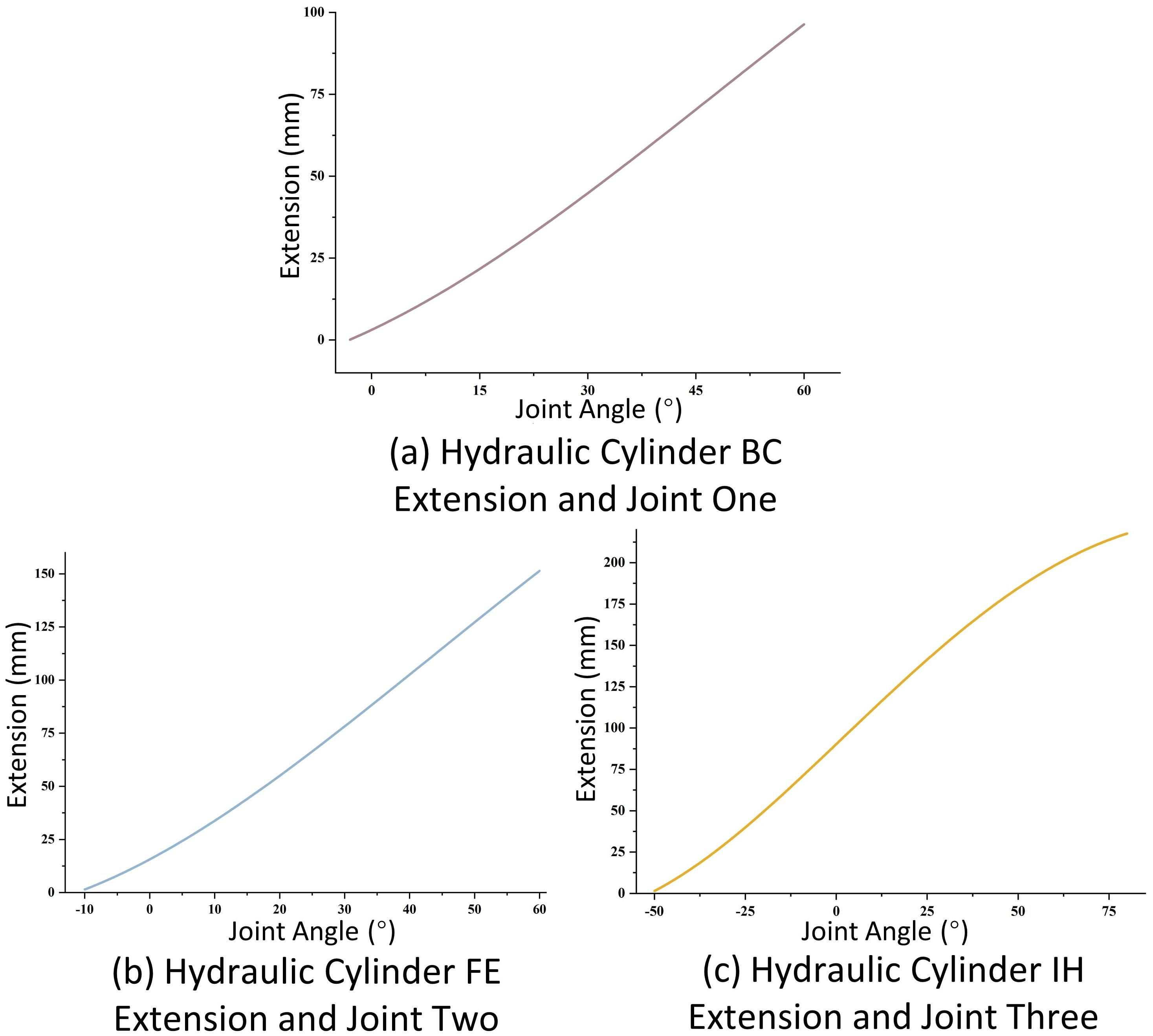}
      \caption{Mapping Relationship Between Hydraulic Cylinder Extension and Joint Angles}
      \label{figurelabel}
   \end{figure}
The mapping relationship between the extension of hydraulic cylinder \(FE\) and joint two, as well as the extension of hydraulic cylinder \(IH\) and joint three, are shown in Figures 5(b) and 5(c).
Assume the forces exerted by hydraulic cylinders \(BC\), \(EF\), and \(HI\) are \(F_1\), \(F_2\), and \(F_3\), respectively. The corresponding output torques are \(T_1\), \(T_2\), and \(T_3\), respectively.
\begin{equation}
\label{deqn_ex1}
T_1=\sin(\pi-\angle ABC)\cdot F_1\cdot l_{AB}
\end{equation}
\begin{equation}
\label{deqn_ex1}
T_2=\sin(\pi-\angle DEF)\cdot F_2\cdot l_{DE}
\end{equation}
\begin{equation}
\label{deqn_ex1}
T_3=\sin(\pi-\angle HIG)\cdot F_3\cdot l_{IG}
\end{equation}

\subsection{Kinematic Model of the Foot}
The hexapod mobile chassis comprises 18 degrees of freedom (DOF) , providing excellent adaptability to complex terrains. Each leg consists of a coxa, femur, and tibia, contributing to 3 DOFs per leg. There is a rotational joint between the coxa and the body, between the coxa and the femur, and between the femur and the tibia, each driven by a hydraulic cylinder.
\begin{figure}[htb]
      \centering
      \includegraphics[width=\columnwidth]{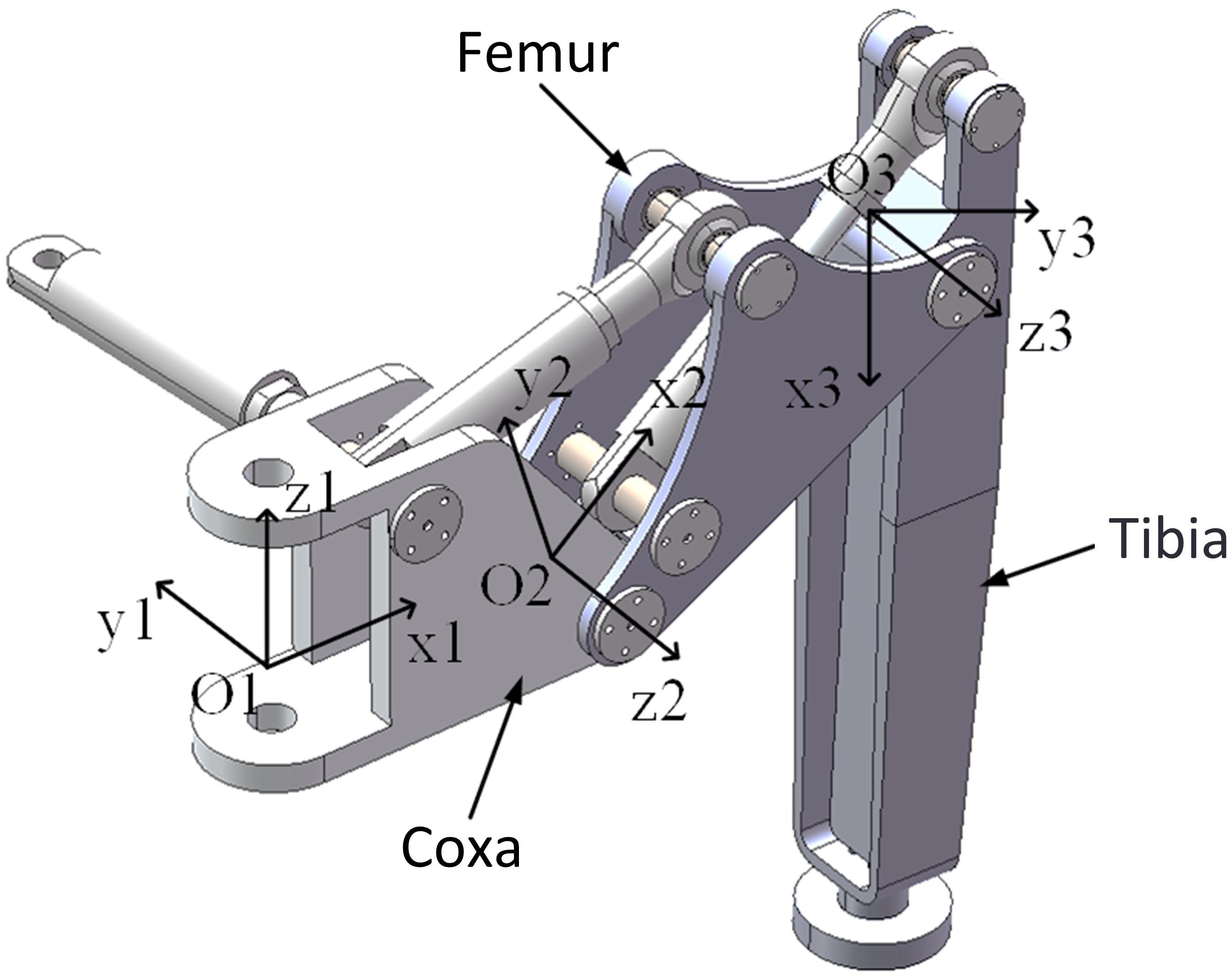}
      \caption{Model of the Robotic Foot}
      \label{figurelabel}
   \end{figure}
   
Taking the right front leg of the robot as an example, the D-H parameters are used to establish the single-leg D-H coordinate system as shown in Figure 6. In this single-leg D-H coordinate system, the "body-coxa" joint, "coxa-femur" joint, and "femur-tibia" joint are sequentially defined as reference coordinate system \(O_1\), coordinate system \(O_2\), and coordinate system \(O_3\). Here, \(O_i\) represents the origin of each coordinate system, \(z_i\) represents the axis of joint \(i\), \(x_i\) represents the common perpendicular between joint axes \(i\) and \((i+1)\), and the direction of \(y_i\) is determined by the right-hand rule.

The end-effector coordinate system in the single-leg coordinate frame of the robot can be obtained by transforming the body coordinate system using the transformation matrix \({}^B T_L\). The formula is as follows:
\begin{equation}
\label{deqn_ex1}
^BT_L=A_1^0A_2^1A_3^2=
\begin{bmatrix}
^BR_L & ^BP_L \\
O & 1
\end{bmatrix}\in R^{4\times4}
\end{equation}
In the formula, \({}^B R_L\) is the rotation matrix, and \({}^B P_L\) is the translation matrix.

By solving the link transformation matrices \(A_i^{i-1}\) between each joint coordinate system and the pose transformation matrix \({}^B T_L\) from the end-effector coordinate system to the root joint coordinate system, the end-effector coordinates \((P_X, P_Y, P_Z)\) can be calculated. The calculation formula for the link transformation matrix \(A_i^{i-1}\) is given in Equation (13).
\begin{equation}
\label{deqn_ex1}
A_i^{i-1}=
\begin{bmatrix}
\cos\theta_i & -\sin\theta_i\cos\alpha_i & \sin\theta_i\sin\alpha_i & a_i\cos\theta_i \\
\sin\theta_i & \cos\theta_i\cos\alpha_i & -\cos\theta_i\sin\alpha_i & a_i\sin\theta_i \\
0 & \sin\alpha_i & \cos\alpha_i & d_i \\
0 & 0 & 0 & 1
\end{bmatrix}
\end{equation}
In the formula, \(A_i^{i-1}\) represents the position and orientation of the \(i\)-th coordinate system relative to the \((i-1)\)-th coordinate system. By substituting the DH parameters into Equation (13), the link transformation matrices between each joint coordinate system can be calculated, as follows:
\begin{equation}
\label{deqn_ex1}
A_1^0=
\begin{bmatrix}
\cos\theta_1 & 0 & \sin\theta_1 & a_1\cos\theta_1 \\
\sin\theta_1 & 0 & -\cos\theta_1 & a_1\sin\theta_1 \\
0 & 1 & 0 & 0 \\
0 & 0 & 0 & 1
\end{bmatrix}
\end{equation}
\begin{equation}
\label{deqn_ex1}
A_2^1=
\begin{bmatrix}
\cos\theta_2 & -\sin\theta_2 & 0 & a_2\cos\theta_2 \\
\sin\theta_2 & \cos\theta_2 & 0 & a_2\sin\theta_2 \\
0 & 0 & 1 & 0 \\
0 & 0 & 0 & 1
\end{bmatrix}
\end{equation}
\begin{equation}
\label{deqn_ex1}
A_3^2=
\begin{bmatrix}
\cos\theta_3 & -\sin\theta_3 & 0 & a_3\cos\theta_3 \\
\sin\theta_3 & \cos\theta_3 & 0 & a_3\sin\theta_3 \\
0 & 0 & 1 & 0 \\
0 & 0 & 0 & 1
\end{bmatrix}
\end{equation}
 By substituting the link transformation matrices between the joint coordinate systems into Equation (12), the transformation matrix from the end-effector coordinate system to the body coordinate system can be calculated as follows:  
\begin{equation}
\label{deqn_ex1}
^BT_L=A_1^0A_2^1A_3^2=
\begin{bmatrix}
n_x & o_x & a_x & P_X \\
n_y & o_y & a_y & P_Y \\
n_z & o_z & a_z & P_Z \\
0 & 0 & 0 & 1
\end{bmatrix}
\end{equation}
The obtained translation matrix \({}^B P_L\) represents the end-effector coordinates \((P_X, P_Y, P_Z)\) as follows:
\begin{equation}
\label{deqn_ex1}
\begin{bmatrix}
P_X \\
P_Y \\
P_Z
\end{bmatrix}=
\begin{bmatrix}
\cos\theta_1\left(a_3\cos\left(\theta_2+\theta_3\right)+a_2\cos\theta_2+a_1\right) \\
\sin\theta_1\left(a_3\cos\left(\theta_2+\theta_3\right)+a_2\cos\theta_2+a_1\right) \\
a_3\sin\left(\theta_2+\theta_3\right)+a_2\sin\theta_2
\end{bmatrix}
\end{equation}
The above process involves solving the forward kinematics, which calculates the relative coordinates of the end-effector with respect to the body based on known joint rotation angles, thereby obtaining the spatial coordinates of the end-effector for the legged construction robot. Solving the inverse kinematics involves using the known position and orientation of the body and the position coordinates of the end-effector. By employing left-multiplication with the inverse transformation matrix, the joint variables \(\theta_i\) can be isolated, thus obtaining the joint angle information. The formula for left-multiplying with the inverse transformation matrix, based on the previously established forward kinematics equations, is as follows:
\begin{equation}
\label{deqn_ex1}
\left(A_1^0\right)^{-1} \, {}^B T_L=A_2^1A_3^2
\end{equation}
Assuming the known end-effector position information is \((P_X, P_Y, P_Z)\), by solving Equation (19) and equating corresponding elements on both sides of the equation, the joint angles \(\theta_1\), \(\theta_2\), and \(\theta_3\) for the single leg can be obtained as follows:
\begin{equation}
\label{deqn_ex1}
\begin{aligned}
 & 
\begin{cases}
\theta_1=\arctan\left(\frac{P_Y}{P_X}\right) \\
\theta_2=\arcsin\left(\frac{P_Z}{\sqrt{\tau+a_1^2}}\right)+\arccos\left(\frac{\tau+a_1^2+a_2^2-a_3^2}{2a_2\sqrt{\tau+a_1^2}}\right) \\
\theta_3=-\arccos\left(\frac{\tau+a_1^2-a_2^2-a_3^2}{2a_2a_3}\right) & 
\end{cases}
\end{aligned}
\end{equation}
In the formula, \(\tau = P_X^2 + P_Y^2 + P_Z^2 - 2a_1 \sqrt{P_X^2 + P_Y^2}\); \(a_1\), \(a_2\), and \(a_3\) are the link lengths of the coxa, femur, and tibia, respectively. 
\section{COORDINATED LEG-ARM MOTION PLANNING}
The hardware system for the coordinated leg-arm motion planning is shown in Figure 7, primarily comprising six legs, a folding arm, and a serial-parallel manipulator. The system includes a total of 30 hydraulic cylinders. The serial-parallel manipulator consists of three modules, each containing two active cylinders and one passive cylinder. The closed-loop control system includes an industrial computer, motion controller, amplifier board, hydraulic cylinders, and sensors.
\begin{figure}[htb]
      \centering
      \includegraphics[width=\columnwidth]{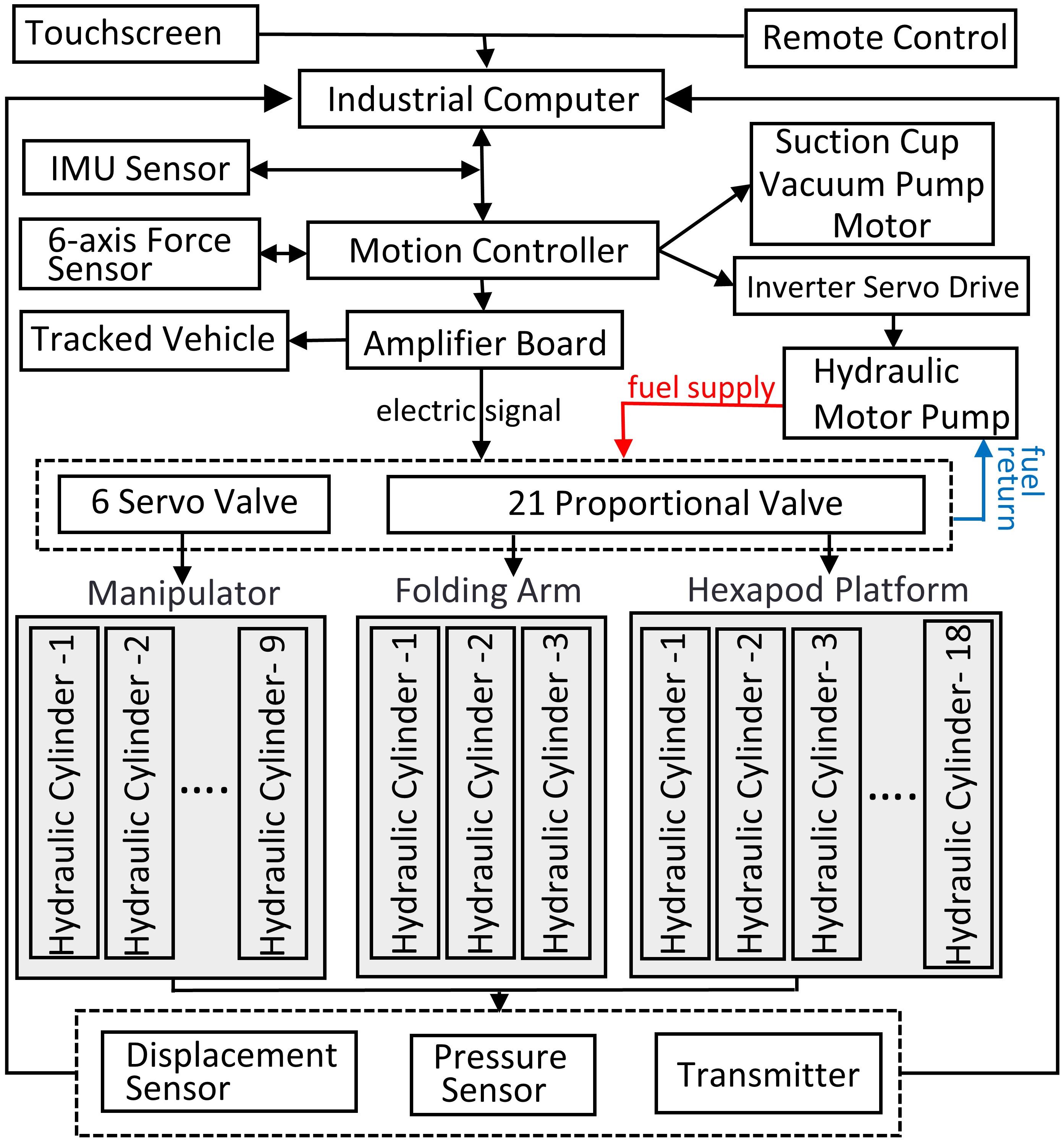}
      \caption{Schematic Diagram of the Hardware Control System}
      \label{figurelabel}
   \end{figure}
   
This study developed a human-machine interaction software using PyQt5, which includes a user interface, an automatic interface, a manual interface, and a debugging interface. Figure 8 shows the human-machine interaction interface of the operation software, featuring a collaborative arm-leg operation module specifically designed for debugging cooperative tasks. This module supports cooperative installation, cooperative grasping, and configuring various other arm-leg cooperative modes.
\begin{figure}[htb]
      \centering
      \includegraphics[width=\columnwidth]{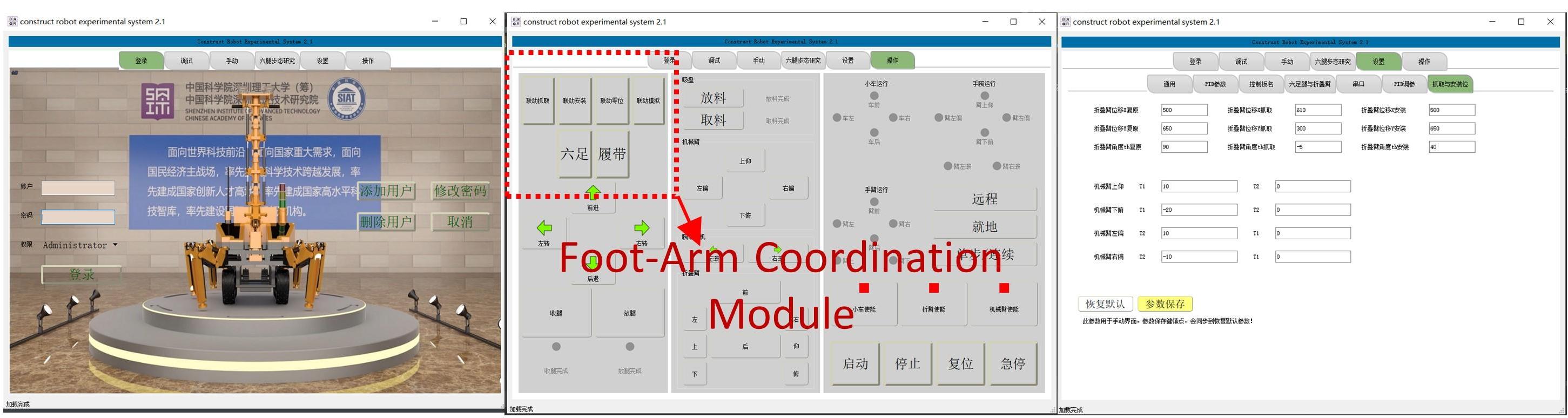}
      \caption{Human-Machine Interaction Interface}
      \label{figurelabel}
   \end{figure}
   
We propose a hierarchical optimization-based whole-body control framework for the robot, achieving global torque optimization while ensuring multiple task completion through coordinated leg-arm motion planning, as illustrated in Figure 9. The high-level planner decomposes the overall motion into sub-motions that meet the requirements of arm-leg coordination. Specifically, the high-level planner performs comprehensive planning, including trunk motion, end-effector motion, and swinging leg motion, to address leg-arm collaboration issues. It also handles the motion planning for folding arms, serial-parallel arms, and hexapod legs, and conveys the results in mathematical form to the lower-level controller.

The low-level controller employs a multi-objective optimization algorithm to constrain the whole-body torque through task-specific constraints, achieving optimal joint torques that meet these requirements. It decomposes the overall motion into sub-motions that satisfy arm-leg coordination needs, thereby enabling comprehensive whole-body control of the leg-arm collaborative robot.
\begin{figure}[htb]
      \centering
      \includegraphics[width=\columnwidth]{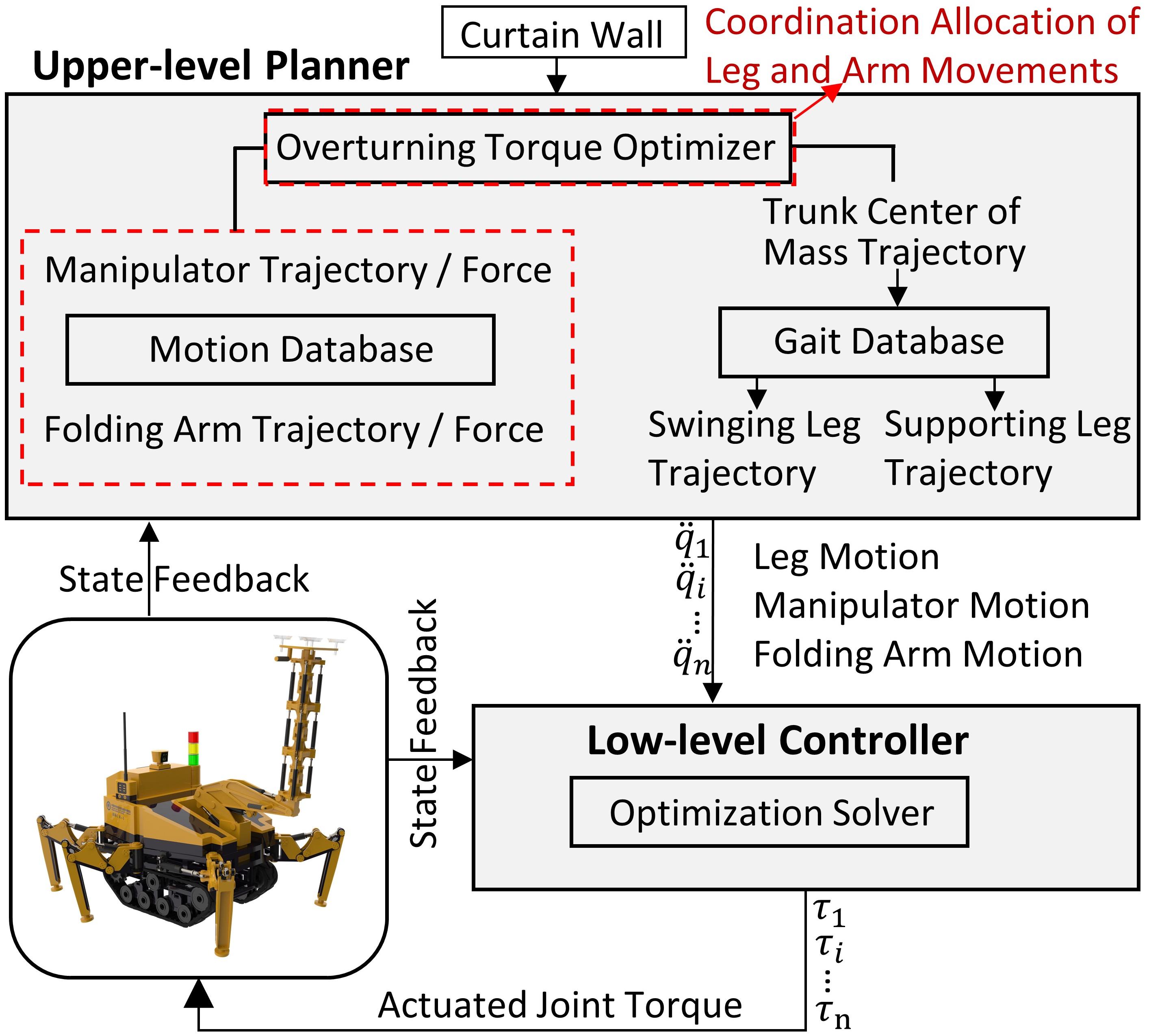}
      \caption{Whole-Body Control Framework Based on Task Space}
      \label{figurelabel}
   \end{figure}
\section{COORDINATED LEG-ARM EXPERIMENT}
\subsection{Fixed-point Turning Experiment}
The robot is capable of performing installation tasks within constrained spaces, demonstrating flexibility and operability. To assess the robot's flexibility, we conducted a fixed-point turning experiment, as shown in Figure 10. In this experiment, the robot successfully executed a 360° rotation, demonstrating its ability to adapt to narrow environments and providing support for operations in confined spaces.
\begin{figure}[htb]
      \centering
      \includegraphics[width=\columnwidth]{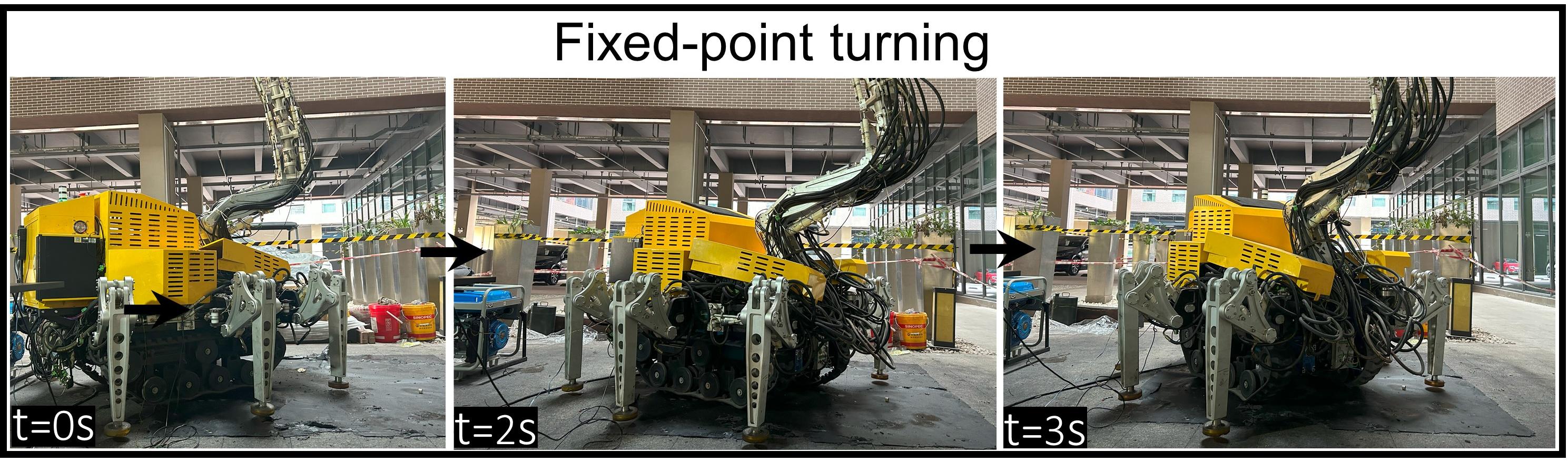}
      \caption{Experiment on Fixed-point Turning}
      \label{figurelabel}
   \end{figure}
\subsection{Simultaneous Installation and Walking Experiment}
First, the robot uses suction cups to grasp the curtain wall panels from the ground. For precise docking and correct alignment during installation at the designated position, adjustments to the pose of the robotic arm and its end-effector are necessary, along with coordinated movement of the hexapod legs.

To assess the multi-task coordination capabilities of the hexapod curtain wall installation robot in dynamic environments, we conducted two sets of experiments to validate the effectiveness of whole-body control for coordinated leg-arm planning. These experiments simulated construction scenarios, demonstrating simultaneous movement and installation, as well as simultaneous movement and pose adjustment.

Experiment 1: Simultaneous Walking and Installation. As shown in Figure 11(a), in the first experiment, the robot demonstrated its capability to perform curtain wall installation while continuously moving. Utilizing whole-body control motion planning, the robot coordinated leg movement and arm operation in real-time during forward motion. The hexapod legs adjusted their posture dynamically to maintain balance and stability, while the robotic arm accurately positioned the curtain wall panels according to pre-defined paths and commands.

Experiment 2: Simultaneous Walking and Adjustment. During actual construction, curtain wall panels frequently require angle and position adjustments to accommodate different building structures. As shown in Figure 11(b), in the second experiment, the robot demonstrated its capability to dynamically adjust the installation pose while continuously moving. The robot performed real-time dynamic corrections of the installation angle during movement, ensuring precise alignment with the building structure.
\begin{figure}[htb]
      \centering
      \includegraphics[width=\columnwidth]{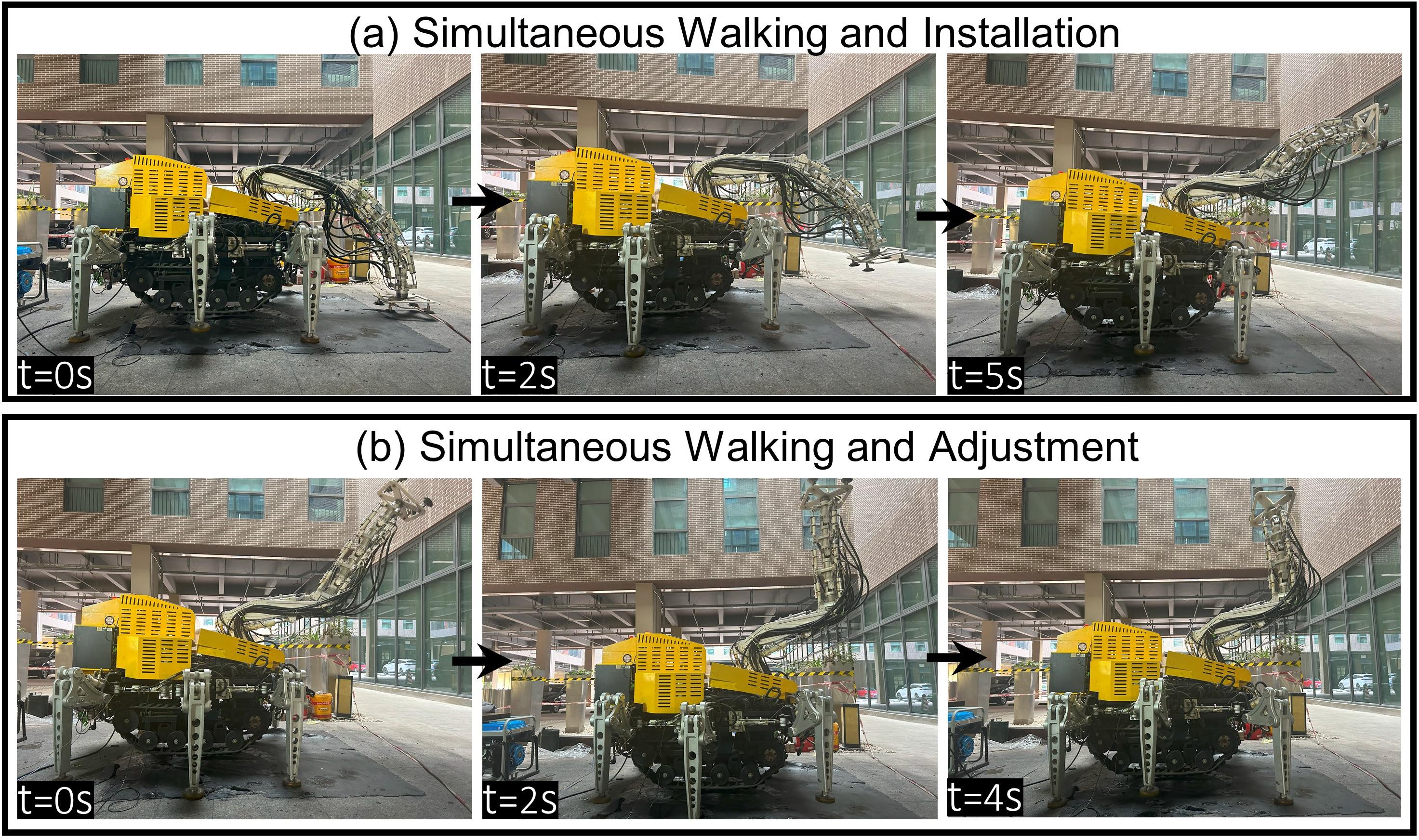}
      \caption{Simultaneous Walking and Installation Experiment}
      \label{figurelabel}
   \end{figure}

The robot can perform the actions of gripping and installing curtain walls, meeting the required installation angles. The rotational and pitch adjustments of the robotic arm and its end-effector exhibited high flexibility and stability. During hexapod locomotion, the gait adjustments were flexible, with no slipping or falling incidents observed. Furthermore, the experiments verified the robot's adaptability to various angle requirements. The results demonstrate that the robot can perform stable operations and pose adjustments while moving, confirming the effectiveness of the whole-body control and coordinated leg-arm planning strategy. This significantly improves construction efficiency and safety, providing reliable technical support for automated curtain wall installation in complex environments.
\section{CONCLUSIONS}

This study focuses on three core operational scenarios in curtain wall installation: wall, ceiling, and floor laying. Based on a hexapod-folding arm-serial-parallel manipulator curtain wall installation robot, we developed a hierarchical optimization-based whole-body control framework for coordinated leg-arm planning to unify the control of hexapod locomotion and arm operations, ensuring efficient task execution and overall torque optimization. Through a series of experiments, including simultaneous walking and installation, and simultaneous walking and adjustment, we validated the effectiveness and superior performance of the control method. Experimental results show that the robot can smoothly perform coordinated leg-arm tasks, including completing curtain wall installations and dynamically adjusting installation poses while continuously moving, thereby meeting complex construction requirements. This significantly improves work efficiency and reduces construction time. Our findings not only demonstrate the engineering effectiveness of the hexapod-folding arm-serial-parallel manipulator framework but also provide solid technical and theoretical support for new operational modes of mobile manipulation robots in construction automation.

\addtolength{\textheight}{-12cm}   




\section*{ACKNOWLEDGMENT}

This work was supported in part by the National Key R\&D Program of China (No.2023YFB4705002), in part by the National Natural Science Foundation of China(U20A20283), in part by the Guangdong Provincial Key Laboratory of Construction Robotics and Intelligent Construction (2022KSYS 013), in part by the CAS Science and Technology Service Network Plan (STS) - Huangpu Special Project (No. STS-HP-202302), in part by the Science and Technology Cooperation Project of Chinese Academy of Sciences in Hubei Province Construction 2023.

\end{document}